\documentclass[runningheads]{llncs}
\usepackage{graphicx}
\usepackage{comment}
\usepackage{amsmath,amssymb} 
\usepackage{color}
\usepackage{wrapfig}
\usepackage{tabularx}
\usepackage{multirow, booktabs}
\usepackage{siunitx}
\usepackage{hyperref}

\newcommand{\titlecap}[2]{\textbf{#1} #2}

\usepackage{xspace}
\makeatletter
\DeclareRobustCommand\onedot{\futurelet\@let@token\@onedot}
\def\@onedot{\ifx\@let@token.\else.\null\fi\xspace}
\def\eg{\emph{e.g}\onedot} 
\def\ie{\emph{i.e}\onedot}

\def\etal{\emph{et al}\onedot}
\makeatother

\newcommand{\Ptpc}{\emph{PT2PC}\xspace}
\newcommand{\Ptpcfull}{``part tree''-to-``point cloud''\xspace}

\begin{document}
\pagestyle{headings}
\mainmatter
\def\ECCVSubNumber{xxx}  

\title{PT2PC: Learning to Generate 3D Point Cloud Shapes from Part Tree Conditions} 

\titlerunning{PT2PC: Part-Tree to Point-Cloud Generation}
\author{Kaichun Mo\inst{1}
\and
He Wang\inst{1} \and
Xinchen Yan\inst{2} \and
Leonidas Guibas\inst{1}}
\authorrunning{K. Mo, H. Wang, X. Yan, and L. Guibas}
%
\institute{Stanford University \and Uber ATG}
\maketitle

\begin{abstract}
3D generative shape modeling is a fundamental research area in computer vision and interactive computer graphics, with many real-world applications.
This paper investigates the novel problem of generating a 3D point cloud geometry for a shape from a symbolic part tree representation.
In order to learn such a conditional shape generation procedure in an end-to-end fashion, we propose a conditional GAN \Ptpcfull model (\Ptpc) that disentangles the \textit{structural} and \textit{geometric} factors.
The proposed model incorporates the part tree condition into the architecture design by passing messages \textit{top-down} and \textit{bottom-up} along the part tree hierarchy.
Experimental results and user study demonstrate the strengths of our method in generating perceptually plausible and diverse 3D point clouds, given the part tree condition.
We also propose a novel structural measure for evaluating if the generated shape point clouds satisfy the part tree conditions.
%
Code and data are released on the webpage: 
{\color{blue}\underline{\url{https://cs.stanford.edu/~kaichun/pt2pc}}}.

\end{abstract}

\section{Introduction}
\label{sec:intro}
%
%
3D shape generation is a central topic in computer vision and graphics. 
Recent works (\eg \cite{choy20163d,fan2017point,gkioxari2019mesh}) have been focusing on generating the entire shape geometry without explicitly considering part semantics and shape structures.
Such holistic shape generation pipelines, though successfully learning to model simple 3D shapes, usually have a difficult time modeling complicated shape structures and delicate shape parts.
In computer-aided design (CAD), constructing a whole shape geometry from scratch is an extremely laborious and time-consuming task.
If the designer only needs to give a sentence “a chair with 1 seat, 4 legs and a back with 3 bars” and the system can directly generate multiple shape candidates for her to select and edit from, it will save a big amount of time.
%
%
%
Disentangling shape structure and geometry factors in shape generation also encourages more fine-grained and controllable 3D shape generation -- thus supporting many real-world applications, including structure-conditioned shape design~\cite{Mo_2019_CVPR,mo2019structurenet} and structure-aware shape re-synthesis~\cite{funkhouser2004modeling}.
%
%
%
%

\begin{figure}[t]
\centering
  \includegraphics[width=0.8\linewidth]{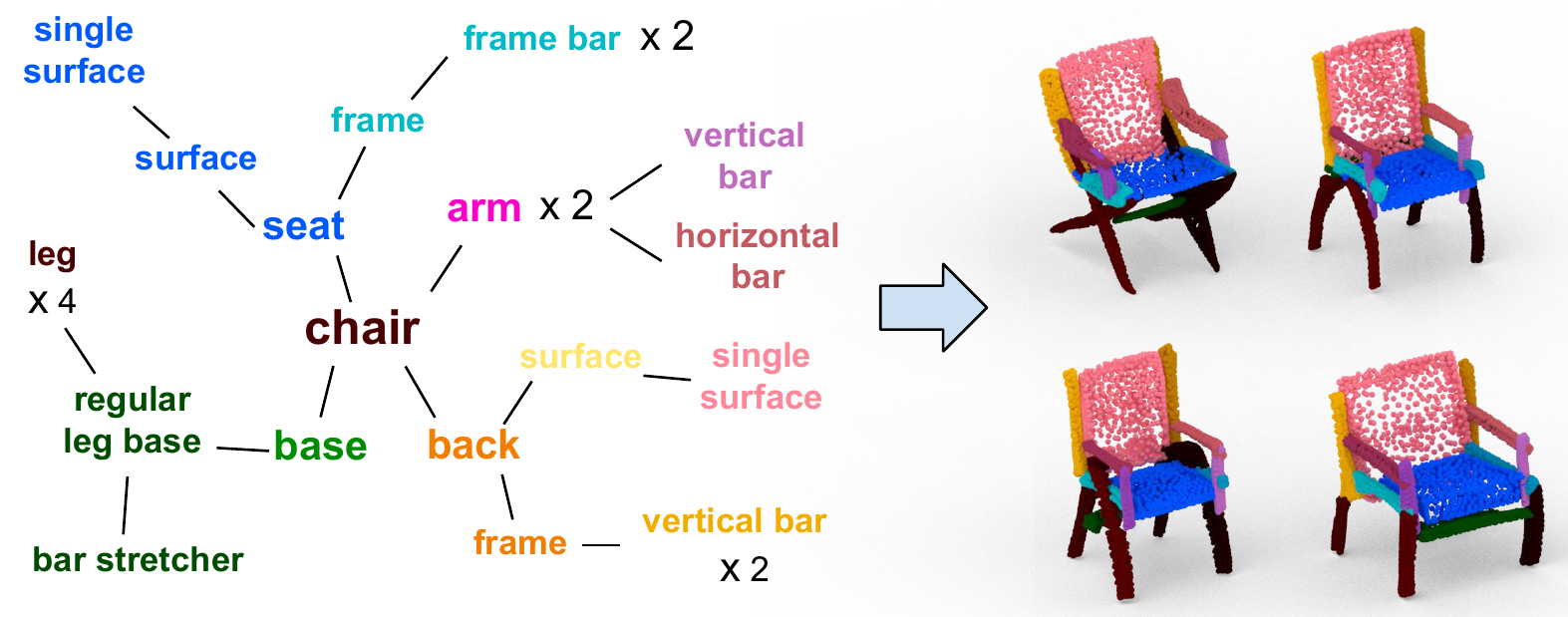}
  \caption{We formulate the problem of \Ptpcfull (\Ptpc) synthesis as a conditional generation task which takes in a symbolic part tree as condition and generates multiple diverse point clouds satisfying the structure defined by the part tree.}
  \label{fig:task_def}
\end{figure}

%
In this paper, we formulate a new task of generating 3D point cloud shapes with \emph{geometric} variations conditioned on \textit{structural} shape descriptions.
Figure~\ref{fig:task_def} illustrates our task input and output with an example.
More specifically, we represent each 3D shape as a hierarchy of parts, following PartNet~\cite{Mo_2019_CVPR}, where each part node has an associated semantic label and the part hierarchy includes parts at different segmentation granularities.
Abstracting away the concrete part geometry, the shape structure can be defined by a symbolic part tree with only part semantics and their relationships (Figure~\ref{fig:task_def}, left).
Given such symbolic part tree conditions, we propose a conditional-GAN \Ptpc to generate diverse 3D point cloud shapes that satisfy the structural conditions (Figure~\ref{fig:task_def}, right).

The symbolic part tree conditions are central to the architecture designs for both \Ptpc generator and discriminator.
Our generator first encodes the part tree template feature using semantic and structural information for each part node in a \emph{bottom-up} fashion along the tree hierarchy.
Then, given a random noise vector capturing the global geometry information at the root node, we recursively propagate such geometric information to each part node in a \emph{top-down} fashion along the part tree. 
The final point clouds are generated by aggregating the point clouds decoded at each leaf node representing its corresponding fine-grained semantic part.
Our discriminator first computes per-part features at the leaf level, propagates the information in a \emph{bottom-up} fashion along the tree hierarchy until the root node and finally produces a score judging if the generated shape \emph{geometry} looks realistic and the shape \emph{structure} satisfies the input condition.
%
%
%
%

%
%

We evaluate the proposed model on four major semantic classes from the PartNet dataset.
To justify the merits of our tree-structure architecture for both the generator and discriminator, we compare with two conditional GAN baselines.
Both quantitative and qualitative results demonstrate clear advantages of our design in terms of global shape quality, part shape quality, and shape diversity, under both seen and unseen templates as the condition.
Results on human evaluation agree with our observations in the experiments and further strengthen our claims.
Additionally, 
we propose a novel hierarchical part instance segmentation method that is able to segment an input point cloud without any part labels into a symbolic part tree. 
This provides us a metric to evaluate how well our generated shape geometry satisfies the part tree conditions.

In summary, our contributions are:
\begin{itemize}
\item we formulate the novel task of part-tree conditioned point cloud generation;
\item we propose a conditional GAN method, \emph{PT2PC}, that generates realistic and diverse point cloud shapes given symbolic part tree conditions;
\item we demonstrate superior performance both quantitatively and qualitatively under standard GAN metrics and a user study, comparing against two baseline conditional-GAN methods;
\item we propose a novel point cloud structural evaluation metric for evaluating if the generated point clouds satisfy the part tree conditions.
\end{itemize}


\section{Related Works}
\label{sec:related}
We review related works on 3D generative shape modeling, part-based shape modeling and structure-conditioned content generation.

\paragraph{3D Generative Shape Modeling.}
Reconstructing and synthesizing 3D shapes is a popular research topic in computer vision and graphics.
Recently, tremendous progresses have been made in generating 
3D voxel grids~\cite{choy20163d,girdhar2016learning,wu2017marrnet,wu2016learning,wu20153d,yan2016perspective,riegler2017octnet}, point clouds~\cite{achlioptas2018learning,fan2017point,gadelha2018multiresolution,yang2018foldingnet,yang2019pointflow}, and surface meshes~\cite{sinha2017surfnet,groueix2018papier,gkioxari2019mesh,liao2018deep} using deep neural networks.
%
%
%
%
Point clouds representation is a collection of unordered points irregularly distributed in the 3D space, which makes the minimax optimization very challenging~\cite{li2018point,achlioptas2018learning}.
Achlioptas~\etal~\cite{achlioptas2018learning} proposed a latent-GAN approach that conducts minimax optimization on the shape feature space which outperforms the raw-GAN operating on the raw point clouds.
To better capture the local geometric structure of point clouds, Valsesia~\etal~\cite{valsesia2018learning} proposed a graph-based generator that dynamically builds the graph based on distance in feature space.
Shu~\etal~\cite{shu20193d} proposed Tree-GAN with a tree-structured graph convolutional neural network as the generator.
Recently, Wang~\etal~\cite{wang2020rethinking} proposed a new discriminator, PointNet-Mix, that improves the sampling uniformity of the generated point clouds.
Unlike these shape point cloud GAN works that generate shapes without explicit part semantic and structural constraints, 
we learn to generate diverse point cloud shapes satisfying symbolic part tree conditions.
%

\paragraph{Part-based Shape Modeling.}
There is a line of research on understanding shapes by their semantic parts and structures.
Previous works study part segmentation~\cite{chen2009benchmark,kalogerakis2010learning,yi2016scalable,kalogerakis20173d,qi2017pointnet,yi2017syncspeccnn,wang20183d,Mo_2019_CVPR,deng2019cvxnets}, box abstraction~\cite{tulsiani2017learning,zou20173d,niu2018im2struct,sun2019learning}, shape template fitting~\cite{kim2013learning,ganapathi2018parsing,genova2019learning,paschalidou2019superquadrics}, generating shapes by parts~\cite{kalogerakis2012probabilistic,li2017grass,sung2017complementme,wang2018global,wu2019sagnet,tian2019learning,mo2019structurenet,wu2019pq,gao2019sdm,schor2019componet,Li:2020}, or editing shape by parts~\cite{fish2014meta,yumer2015semantic,Mo19StructEdit}.
We refer to the survey papers~\cite{xu2016data,mitra2014structure} for more related works.
%
%
Shape parts have hierarchical structures~\cite{wang2011symmetry,van2013co,Mo_2019_CVPR}.
Yi~\etal~\cite{yi2017learning} learns consistent part hierarchy from noisy online tagged shape parts.
GRASS~\cite{li2017grass} propose binary part trees to generate novel shapes.
A follow-up work~\cite{yu2019partnet} learns to segment shapes into the binary part hierarchy.
PartNet~\cite{Mo_2019_CVPR} proposes a large-scale 3D model dataset with hierarchical part segmentation.
Using PartNet, recent works such as StructureNet~\cite{mo2019structurenet} and StructEdit~\cite{Mo19StructEdit} learns to generate and edit shapes explicitly following the pre-defined part hierarchy.
We use the tree hierarchy defined in PartNet~\cite{Mo_2019_CVPR} and propose a new task \emph{PT2PC} that learns to generate point cloud shapes given symbolic part tree conditions.

\paragraph{Conditional Content Generation.}
Understanding the 3D visual world, parsing the geometric and structural properties of 3D primitives (\eg objects in the scene or parts of an object) and their relationships is at the core of computer vision~\cite{gupta2010blocks,xu2017scene,johnson2018image,tulsiani2018factoring}.
Many works learn to synthesize high-quality images from text descriptions~\cite{karacan2016learning,reed2016learning,reed2016generative,zhang2017stackgan,yin2019semantics,li2019object,tan2019text2scene}, semantic attributes~\cite{yan2016attribute2image,choi2018stargan}, scene-graph representations~\cite{johnson2018image,ashual2019specifying}, and rough object layouts~\cite{hong2018inferring,hong2018learning,zhao2019image,mo2018instagan}.
There are also works to generate 3D content with certain input conditions.
Chang~\etal~\cite{chang2014learning,chang2015text} learns to generate 3D scenes from text.
Chen~\etal~\cite{chen2018text2shape} studied how to generate 3D voxel shapes from a sentence condition.
StructEdit~\cite{Mo19StructEdit} learns to generate structural shape variations conditioned on an input source shape.
Our work introduces a conditional Generative Adversarial Network that generates shape point clouds conditioned on an input symbolic part tree structure.

%



\section{Method}
\label{sec:method}
In this work, we propose \Ptpc, a conditional GAN (c-GAN) that learns a mapping from a given \emph{symbolic part tree} $\mathcal{T}$ and a random noise vector $\mathbf{z}$ to a 3D shape point cloud $\mathbf{X}$ composed of part point clouds for the leaf nodes of the conditional part tree.
%
%
We propose novel part-based conditional point cloud  generator $G(\mathbf{z}, \mathcal{T})$ and discriminator $D(\mathbf{X}, \mathcal{T})$ conditioned on the symbolic part tree input $\mathcal{T}$.
Different from holistic point cloud GANs~\cite{achlioptas2018learning,valsesia2018learning,shu20193d} that produce a shape point cloud as a whole, our proposed \Ptpc generate a hierarchy of part point clouds along with part semantics and shape structures.

%
\subsection{Symbolic Part Tree Representation}

\begin{wrapfigure}{r}{0.4\textwidth}
\centering
  \includegraphics[width=\linewidth]{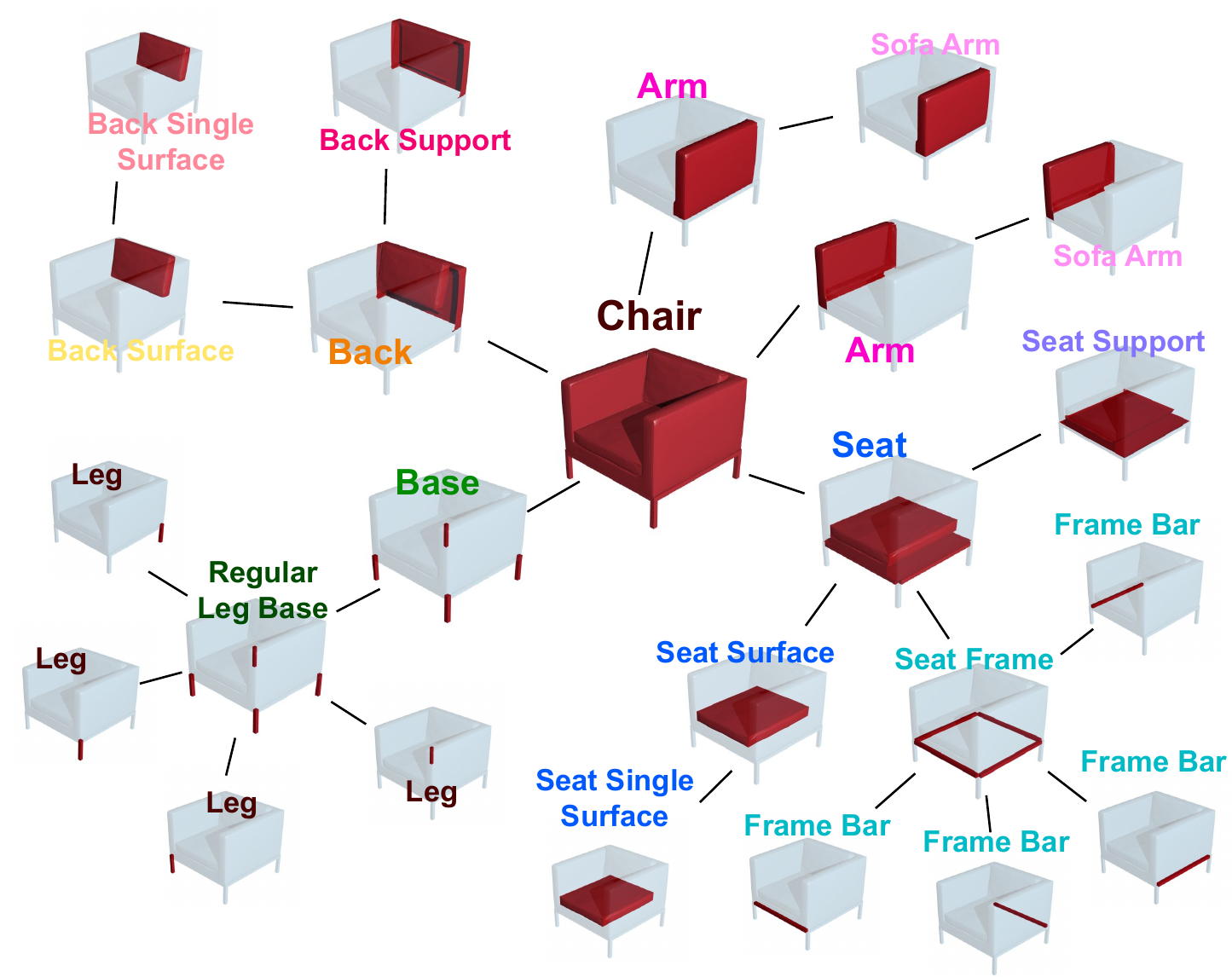}
  \caption{An example PartNet hierarchical part segmentation.}
  \label{fig:part_hier}
\end{wrapfigure}
We follow the semantic part hierarchy defined in PartNet~\cite{Mo_2019_CVPR}.
Every PartNet shape instance (\eg a chair) is annotated with a hierarchical part segmentation that provides both coarse-level parts (\eg chair base, chair back) and parts at fine-grained levels (\eg chair leg, chair back bar).
Figure~\ref{fig:part_hier} shows the ground-truth part hierarchy of an exemplar chair.

A symbolic part tree $\mathcal{T}$ is defined as $\mathcal{T}=(\mathcal{T}_V, \mathcal{T}_E)$, where $\mathcal{T}_V=\{P^j|P^j=(\mathbf{s}^j, \mathbf{d}^j)\}_j$ represents a set of part instances and $\mathcal{T}_E$ represents an directed edge set of the part parent-children relationships $\mathcal{T}_E=\{(j, k)\}$.
In $\mathcal{T}_V$, each part instance $P^j$ is composed of two components: a semantic label $\mathbf{s}^j$ (\eg chair seat, chair back), and a part instance identifier $\mathbf{d}^j$ (\eg the first leg, the second leg), both of which are represented as one-hot vectors.
%
The set of part semantic labels are pre-defined in PartNet and consistent within one object category.
In $\mathcal{T}_E$, each edge $(j, k)$ indicates $P^j$ is the parent node of $P^k$.
The set $C^j=\left\{k|(j, k)\in\mathcal{T}_E\right\}$ defines all children part instances of a node $P^j$.
We denote a special part node $P^\texttt{root}$ to be the root node of the part tree $\mathcal{T}$, with the semantic label $\mathbf{s}^\texttt{root}$ and the instance identifier $\mathbf{d}^\texttt{root}$.
%
The leaf node of the symbolic part tree has no children, namely, $\mathcal{T}_\texttt{leaf}=\left\{P^j\mid|C^j|=0\right\}\subsetneq\mathcal{T}_V$.
%

\begin{figure}[t]
\centering
  \includegraphics[width=\linewidth]{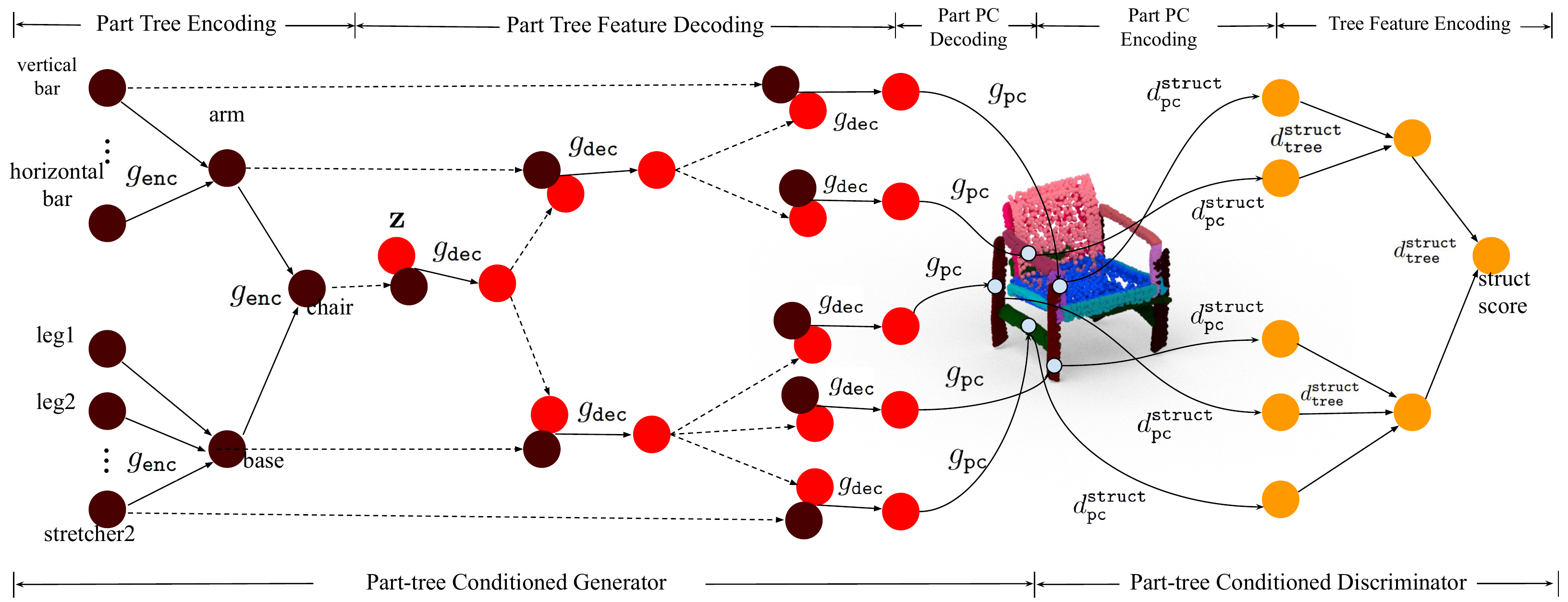}
  \caption{\titlecap{Our c-GAN \emph{PT2PC} architecture.}{Our part-tree conditioned generator first extracts the subtree features by traversing the input symbolic part tree in a \emph{bottom-up} fashion, and then recursively decodes part features in a \emph{top-down} way until leaf nodes, where part point clouds are finally generated. Our part-tree conditioned geometry discriminator recursively consumes the generated part tree with leaf node geometry in a \emph{bottom-up} fashion to generate the final score. The solid arrow indicates a network module while a dashed arrow means to copy the content. As defined in Sec 3, the brown, red and orange nodes represent the encoded symbolic part feature $\mathbf{t}$, the decoded part geometry feature $\mathbf{f}$ and the encoded part geometry feature $\mathbf{h}$ respectively.}}
  \label{fig:arch}
\end{figure}

\subsection{Part-tree Conditioned Generator}
Our conditional generator $G(\mathbf{z},\mathcal{T})$ takes a random Gaussian variable $\mathbf{z} \sim \mathcal{N}(\cdot| \mathbf{\mu}=0, \mathbf{\sigma}=1)$ and a symbolic part tree condition $\mathcal{T}=(\mathcal{T}_V, \mathcal{T}_E)$ as inputs and outputs a set of part point clouds $\mathbf{X}=\{\mathbf{x}^j\mid P^j\in \mathcal{T}_\texttt{leaf}\}$ where $\mathbf{x}^j\in\mathbb{R}^{M\times3}$ is a part point cloud in the shape space representing the leaf node part $P^j$. Namely,
\begin{equation}
    \mathbf{X} = G(\mathbf{z}, \mathcal{T})
\end{equation}

The generator is composed of three network modules: a symbolic part tree encoder $G_\texttt{enc}$, a part tree feature decoder $G_\texttt{dec}$ and a part point cloud decoder $G_\texttt{pc}$.
First, the symbolic part tree encoder $G_\texttt{enc}$ embeds the nodes of $\mathcal{T}$ into compact features $\mathbf{t}^j$ hierarchically from the leaf nodes to the root node for every node $P^j$.
Then, taking in both the random variable $\mathbf{z}$ and the hierarchy of symbolic part features $\{\mathbf{t}^j\}_j$, the part tree feature decoder $G_\texttt{dec}$ hierarchically decodes the part features in the top-down manner, from the root node to the leaf nodes, and finally produces part feature $\mathbf{f}^j$ for every leaf node $P^j\in\mathcal{T}_\texttt{leaf}$.
Finally, the part point cloud decoder $G_\texttt{pc}$ transforms the leaf node features $\left\{\mathbf{f}^j\mid P^j\in\mathcal{T}_\texttt{leaf}\right\}$ into 3D part point clouds $\left\{\mathbf{x}^j\mid P^j\in\mathcal{T}_\texttt{leaf}\right\}$ in the shape space.

At each step of the part feature decoding, the parent node needs to know the global structural context in order to propagate coherent signals to all of its children so that the generated part point clouds can form a valid shape in a compatible way. This is the reason why we introduce the symbolic part tree encoder $G_\texttt{enc}$ as a \emph{bottom-up} module to summarize the part tree structural context for each decoding step. 
Our part tree decoder $G_\texttt{dec}$ is then conditioned on the symbolic structural context and recursively propagates the random noise $\mathbf{z}$ from the root node to the leaf nodes in a \emph{top-down} fashion.

\paragraph{Symbolic part tree encoder $G_\texttt{enc}$.} 
For a given symbolic part tree $\mathcal{T}$, we encode the nodes of the part tree starting from the leaf nodes and propagate the messages to the parent node of the encoded nodes until the root node gets encoded. The message propagation is performed in a \textit{bottom-up} fashion.
As shown in Eq.~\ref{eqn:g_template_encoder},
each node $P^j$ takes the node feature $\mathbf{t}^k$, the semantic label $\mathbf{s}^k$ and the part instance identifier $\mathbf{d}^k$ from all its children $\left\{P^k|k\in C^j\right\}$, aggregates the information and computes its node feature $\mathbf{t}^j$ using a learned function $g_{enc}$. Then, it further propagates a message to its parent node.
We initialize $\mathbf{t}^j=0$ for every leaf node $P^j\in \mathcal{T}_\texttt{leaf}$.
\begin{align}
    \label{eqn:g_template_encoder}
    \textbf{t}^j &= \mathbf{0}, &\forall& P^j\in \mathcal{T}_\texttt{leaf}\nonumber \\
    \textbf{t}^j &= g_\texttt{enc}\left(\left\{[\textbf{t}^k; \textbf{s}^k; \textbf{d}^k]\mid k \in C^j\right\}\right), &\forall& P^j\in \mathcal{T}_V-\mathcal{T}_\texttt{leaf}
\end{align}
where $[\cdot; \cdot]$ means a concatenation of the inputs.
$g_\texttt{enc}$ is implemented as a small PointNet\cite{qi2017pointnet}, treating each children-node feature as a high dimensional point, to enforce the permutation invariance between children nodes. We first use a fully-connected layer to embed each $[\textbf{t}^k; \textbf{s}^k; \textbf{d}^k]$ into a $256$-dim feature, then perform a max-pooling over $K=|C^j|$ features over all children nodes to obtain an aggregated feature, and finally push the aggregated feature through another fully-connected layer to obtain the final parent node feature $\mathbf{t}^j$.
We use leaky \texttt{ReLU} as the activation functions in our fully-connected layers.

\paragraph{Part tree feature decoder $G_\texttt{dec}$.} 
Taking in the random variable $\mathbf{z}$ and encoded node features $\{\mathbf{t}^j\}_j$, we hierarchically decode the part features $\{\mathbf{f}^j\}_j$ from the root node to the leaf nodes in a \textit{top-down} fashion along the given part tree structure $\mathcal{T}$.
As shown in Eq.~\ref{eqn:g_feature_decoder}, for every part $P^j$, we learn a shared function $g_{dec}$ transforming the concatenation of its own features ($\textbf{t}^j, \textbf{s}^j, \textbf{d}^j$) and the decoded feature $\textbf{f}^p$ from its parent node $P^p$ into part feature $\textbf{f}^j$. For the root node, we use random noise $\textbf{z}$ to replace parent node feature.
%
%
\begin{align}
    \label{eqn:g_feature_decoder}
    \textbf{f}^\texttt{root}&= g_\texttt{dec}\left([\textbf{z}; \textbf{t}^\texttt{root}; \textbf{s}^\texttt{root}; \textbf{d}^\texttt{root}]\right)\nonumber, &\\
    \textbf{f}^j &= g_\texttt{dec}\left([\textbf{f}^p; \textbf{t}^j; \textbf{s}^j; \textbf{d}^j]\right),&\forall (p, j)\in \mathcal{T}_E
\end{align}
We implement $g_\texttt{dec}$ as a two-layer MLP with leaky \texttt{ReLU} as the activation functions. The output feature size is 256.

\paragraph{Part point cloud decoder $G_\texttt{pc}$.} Given the part features of all the leaf nodes $\left\{\textbf{f}^j\mid P^j\in\mathcal{T}_\texttt{leaf}\right\}$, our point cloud decoder $G_\texttt{pc}$ transforms each individual feature ${\textbf{f}^j}$ into a 3D part point cloud $\mathbf{x}^j$ in the shape space for every $P^j \in\mathcal{T}_\texttt{leaf}$, as shown in Eq.~\ref{eqn:g_pc_decoder}.
To get the final shape point cloud, we down-sample the union of all part point clouds $\left\{\mathbf{x}^j\mid P^j\in\mathcal{T}_\texttt{leaf}\right\}$.
We generate the same number $M$ points for all the parts. 
%
\begin{align}
\label{eqn:g_pc_decoder}
    \textbf{x}^j &= g_\texttt{pc}(\textbf{f}^j),\forall P^j\in\mathcal{T}_\texttt{leaf} \nonumber \\
    \textbf{x} &= \texttt{DownSample}\left(\cup \left\{\mathbf{x}^j\mid P^j\in\mathcal{T}_\texttt{leaf}\right\} \right)
\end{align}
$g_\texttt{pc}$ is designed to deform a fixed surface point cloud of a unit cube $\mathbf{x}_\texttt{cube}$ into our target part point cloud based on its input $\textbf{f}$, inspired by the shape decoder introduced in Groueix~\etal~\cite{groueix2018atlasnet}.
We uniformly sample a 1000-size point cloud from the surface of a unit cube to form $\mathbf{x}_\texttt{cube}\in\mathbb{R}^{1000\times3}$.
Then, for each point in $\mathbf{x}_\texttt{cube}$, we concatenate its XYZ coordinate with the feature $\mathbf{f}$, push it through a $MLP(256+3, 1024, 1024, 3)$ using leaky \texttt{ReLU}, and finally obtain an XYZ coordinate for a point on our target point cloud.
%
%
Finally, we use Furthest Point Sampling (FPS) for our \texttt{Downsample} operation to obtain shape point cloud $\mathbf{x}$.

Compared to existing works~\cite{achlioptas2018learning,valsesia2018learning,shu20193d} that generate shape point clouds as a whole, the key difference here is that our point cloud decoder generates part point clouds for every leaf node in the part tree $\mathcal{T}$ separately, but in a manner aware of the inter-part structure and relationships. Another big advantage is that we get the semantic label of each generated part point cloud. 
Furthermore, we observe that the holistic point cloud generators usually suffer from non-uniform point distribution. The generators tend to allocate way more points to bulky parts (\eg, chair back and chair seat) while only generating sparse points for small parts with thin geometry (\eg, chair wheel, chair back bar).
Since our $G_\texttt{pc}$ generates the same number of points for each part and then performs global down-sampling, we can generate shape point clouds with fine structures and appropriate point density for all the parts. 

\subsection{Part-tree Conditioned Discriminator}
Our conditional discriminator $D(\mathbf{X}, \mathcal{T})$ receives a generated sample or a true data sample, composed of a set of part point clouds $\mathbf{X}=\left\{\mathbf{x}^j\in \mathbb{R}^{M\times3} |P^j\in\mathcal{T}_\texttt{leaf}\right\}$, and outputs a scalar $y \in \mathbb{R}$ based on the tree condition $\mathcal{T}$. Following the WGAN-gp~\cite{arjovsky2017wasserstein,gulrajani2017improved}, $D$ is learned to be a 1-Lipschitz function of $\mathbf{X}$ and its output $y$ depicts the realness of the sample.  


Since the input $\mathbf{X}$ always contains part point clouds for every leaf node part instances in the symbolic part tree $\mathcal{T}$, 
%
our discriminator mainly focus on judging the geometry of each part point clouds along with the whole shape point clouds assembled from the parts. This is to say, the discriminator should tell if each part point cloud is realistic and plausible regarding its part semantics; in addition, the discriminator needs to look at the spatial arrangement of the part point clouds, judge whether it follows a realistic structure specified by the part tree $\mathcal{T}$, \eg connected parts need to contact each other and some parts may exhibit certain kind of symmetry; finally, the discriminator should judge whether the generated part point clouds form a realistic shape point cloud. 

To address the requirements above, our discriminator leverages two modules: a structure-aware part point cloud discriminator $D^\texttt{struct}(\mathbf{X}, \mathcal{T})$, and a holistic shape point cloud discriminator $D^\texttt{whole}(\mathbf{x})$, where $\mathbf{x}=\texttt{DownSample}\left(\cup \mathbf{X}\right)$.
$D^\texttt{struct}(\mathbf{X}, \mathcal{T})$ takes as input the part tree condition $\mathcal{T}$ and the generated set of part point clouds $\mathbf{X}$ and outputs a scalar $y^\texttt{struct} \in \mathbb{R}$ regarding the tree-conditioned generation quality. $D^\texttt{whole}(\mathbf{x}) \in \mathbb{R}$ only takes the down-sampled shape point cloud $\mathbf{x}$ as input and outputs a scalar $y^\texttt{whole}$ regarding the unconditioned shape quality.
%
As shown in Eq.\ref{eqn:discriminator}, the final output of our discriminator $D$ is the sum of the two discriminators.
\begin{align}
    \label{eqn:discriminator}
    y=&y^\texttt{struct}+y^\texttt{whole} \nonumber \\
    y^\texttt{struct}=&D^\texttt{struct}(\mathbf{X}, \mathcal{T}) \nonumber \\
    y^\texttt{whole}=&D^\texttt{whole}(\mathbf{x})
\end{align}

For the structure-aware part point cloud discriminator $D^\texttt{struct}(\mathbf{X}, \mathcal{T})$, we constitute it using  three network components: a part point cloud encoder $D^\texttt{struct}_\texttt{pc}$, a tree-based feature encoder $D^\texttt{struct}_\texttt{tree}$, and a scoring network $D^\texttt{struct}_\texttt{score}$.
First, the point cloud encoder $D^\texttt{struct}_\texttt{pc}$ encodes the part point cloud $\mathbf{x}^j$ into a part feature $\mathbf{h}^j$ for each leaf node $P^j \in \mathcal{T}_\texttt{leaf}$.
Then, taking in the part features at leaf level $\left\{\textbf{h}^j\mid P^j\in\mathcal{T}_\texttt{leaf}\right\}$, the tree-based feature encoder $D^\texttt{struct}_\texttt{tree}$ recursively propagates the part features $\textbf{h}$ along with the part semantics $\textbf{s}$ to the parent nodes starting from the leaf nodes and finally reaching the root node, in a \textit{bottom-up} fashion.
Finally, a scoring function $D^\texttt{struct}_\texttt{score}$ outputs a score $y^\texttt{struct}\in\mathbb{R}$ for the shape generation quality.
%
For the holistic shape point cloud discriminator $D^\texttt{whole}$, 
%
it is simply composed of a PointNet encoder $D^\texttt{whole}_\texttt{pc}$ and a scoring network $D^\texttt{whole}_\texttt{score}$ which outputs a scalar $y^\texttt{whole}\in\mathbb{R}$.

\paragraph{Point cloud encoder $D^\texttt{struct}_\texttt{pc}$ and $D^\texttt{whole}_\texttt{pc}$.}
Both $D^\texttt{struct}_\texttt{pc}$ and $D^\texttt{whole}_\texttt{pc}$ use vanilla PointNet~\cite{qi2017pointnet} architecture without spatial transformer layers or batch normalization layers.
For $D^\texttt{struct}_\texttt{pc}$, we learn a function $d^\texttt{struct}_\texttt{pc}$ to extract a part geometry feature $\mathbf{h}^j$ for each part point cloud $\mathbf{x}^j$.
\begin{align}
\mathbf{h}^j = d^\texttt{struct}_\texttt{pc}(\mathbf{x}^j), \forall P^j\in\mathcal{T}_\texttt{leaf}  
\end{align}
$d^\texttt{struct}_\texttt{pc}$ is implemented as a four-layer $MLP(3, 64, 128, 128, 1024)$ to process each point individually followed by a max-pooling.
Similarly, $D^\texttt{whole}_\texttt{pc}$ takes a shape point cloud $\mathbf{x}$ as input and outputs a global shape feature $\mathbf{h}^\texttt{shape}$.
\begin{align}
\mathbf{h}^\texttt{shape} = D^\texttt{whole}_\texttt{pc}(\mathbf{x})
\end{align}


\paragraph{Tree feature encoder $D^\texttt{struct}_\texttt{tree}$.}
Similar to the symbolic part tree encoder $G_\texttt{dec}$ in the generator, $D^\texttt{struct}_\texttt{tree}$ learns an aggregation function $d_\texttt{tree}^\texttt{struct}$ that transforms features from children nodes into parent node features, as shown in  Eq.~\ref{eqn:d_part_feature_encoder}. By leveraging the tree structure specified by $\mathcal{T}$ in its architecture, the module enforces the structure-awareness of $D^{\texttt{struct}}$. In a \textit{bottom-up} fashion, the features propagate from the leaf level finally to the root yielding $\mathbf{h}^\texttt{root}$, according to Eq.\ref{eqn:d_part_feature_encoder}.
\begin{align}
    \label{eqn:d_part_feature_encoder}
    \textbf{h}^j =& d_\texttt{tree}^\texttt{struct}\left(\left\{[\textbf{h}^k; \textbf{s}^k]\mid k \in C^j\right\}\right), \forall P^j\in \mathcal{T}_V-\mathcal{T}_\texttt{leaf}
\end{align}
To implement $D_\texttt{tree}^\texttt{struct}$, we extract a latent $256$-dim feature after applying a fully-connected layer over each input $[\textbf{h}^k; \textbf{s}^k]$, perform max-pooling over all children nodes and finally push it through another fully-connected layer to obtain $\textbf{h}^j$.
We use the leaky \texttt{ReLU} activation functions for both layers.

Note that the key difference between $D_\texttt{tree}^\texttt{struct}$ and $G_\texttt{enc}$ is that $D_\texttt{tree}^\texttt{struct}$ no longer requires the part instance identifiers $\mathbf{d}^k$ since the children part features $\{\textbf{h}^k\}_k$ for each parent node $P^j$ already encode the part geometry information that are naturally different even for part instances of the same part semantics.

\paragraph{Scoring functions $D^\texttt{part}_\texttt{score}$ and $D^\texttt{whole}_\texttt{score}$.} 
After obtaining the structure-aware root feature $\mathbf{h}^\texttt{root}$ and the holistic PointNet feature $\mathbf{h}^\texttt{shape}$, we compute
\begin{align}
y^\texttt{struct}&=D^\texttt{struct}_\texttt{score}\left(\mathbf{h}^\texttt{root}\right) \nonumber \\
y^\texttt{whole}&=D^\texttt{whole}_\texttt{score}\left(\mathbf{h}^\texttt{shape}\right)
\end{align}
Both scoring functions are implemented as a simple fully-connected layer with no activation function.

\subsection{Training}
We follow WGAN-gp ~\cite{arjovsky2017wasserstein,gulrajani2017improved} for training our \Ptpc conditional generator  $G(\cdot, \mathcal{T})$ and discriminator $D(\cdot, \mathcal{T})$.
The objective function is defined in Eq.~\ref{eqn:wgan_gp}.
%
\begin{equation}
    \label{eqn:wgan_gp}
    \mathcal{L} = \mathbb{E}_{\mathbf{z} \sim \mathcal{Z}}[D(G(\mathbf{z}, \mathcal{T}), \mathcal{T})] - \mathbb{E}_{\mathbf{X} \sim \mathcal{R}}[D(\mathbf{X}, \mathcal{T})]
    + \lambda_{gp}\mathbb{E}_{\mathbf{\hat{X}}} \left[(\| \nabla_{\mathbf{\hat{X}}} D(\mathbf{\hat{X}}, \mathcal{T})\|_2-1)^2\right]
\end{equation}
where we interpolate each pair of corresponding part point clouds from a real set $\mathbf{X}_\texttt{real} =\left\{\mathbf{x}_\texttt{real}^j|P^j\in\mathcal{T}_\texttt{leaf}\right\}$ and a fake set $\mathbf{X}_\texttt{fake}=\left\{\mathbf{x}_\texttt{fake}^j|P^j\in\mathcal{T}_\texttt{leaf}\right\}$ to get $\mathbf{\hat{X}}=\left\{\mathbf{\hat{x}}^j |P^j\in\mathcal{T}_\texttt{leaf}\right\}$, as shown in below:
\begin{equation}
    \mathbf{\hat{x}}^j=\alpha\cdot \mathbf{x}^j_\texttt{real}+(1-\alpha)\cdot\mathbf{x}^j_\texttt{fake}, \forall P^j\in\mathcal{T}_\texttt{leaf},
\end{equation}
where $ \alpha\in(0,1)$ is a random interpolation coefficient always remaining same for all parts.
We iteratively train the generator and discriminator with $n_{critic} = 10$. We choose $\lambda_{gp}=1$, $N=2,048$ and $M=1,000$ in our experiments. 
And, we assume the maximal children per parent node to be 10.

\section{Experiments}
\label{sec:exp}

We evaluate our proposed \emph{PT2PC} on the PartNet~\cite{Mo_2019_CVPR} dataset and compare to two baseline c-GAN methods.
We show superior performance on all standard point cloud GAN metrics.
Besides, we propose a new structural metric evaluating how well the generated point clouds satisfy the input part tree conditions, based on a novel hierarchical instance-level shape part segmentation algorithm.
We also conduct a user study which confirms our strengths over baseline methods.

\subsection{Dataset}

\begin{wraptable}{r}{0.5\textwidth}
\caption{\titlecap{Dataset Statistics.} We summarize the number of shapes and symbolic part trees in the train and test splits for each object category.} 
\centering
\label{tab:dataset}
{\small
\begin{tabularx}{0.5\textwidth}{>{\centering}m{1.3cm}|ccc|ccc}
    \toprule
    \multirow{2}{*}{Category} &  \multicolumn{3}{c|}{\#Shapes} & \multicolumn{3}{c}{\#Part Trees} \\
    & Total & Train & Test  & Total & Train & Test \\
    \midrule
     Chair & 4871 & 3848 & 1023 & 2197 & 1648 & 549 \\
     Table & 5099 & 4146 & 953 & 1267 & 925 & 342 \\
     Cabinet & 846 & 606 & 240 & 619 & 470 & 149 \\
     Lamp & 802 & 569 & 233 & 302 & 224 & 78 \\
    \bottomrule
\end{tabularx}
}
\end{wraptable}
We use the PartNet~\cite{Mo_2019_CVPR} dataset as our main testbed.
PartNet contains fine-grained and hierarchical instance-level semantic part annotations including 573,585 part instances over 26,671 3D models covering 24 categories.
In this paper, we use the four largest categories that contain diverse part structures: chairs, tables, cabinets and lamps.
Following StructureNet~\cite{mo2019structurenet}, we assume there are at maximum 10 children for every parent node and remove the shapes containing unlabeled parts for the canonical sets of part semantics in PartNet~\cite{Mo_2019_CVPR}.
Table~\ref{tab:dataset} summarizes data statistics and the train/test splits.
We split by part trees with a ratio 75\%:25\%. 
See Table~\ref{tab:dataset} for more details.
We observe that most part trees (\eg 1,787 out of 2,197 for chairs) have only one real data point in PartNet, which posts challenges to generate shapes with geometry variations.

\begin{figure}[!htb]
\centering
  \includegraphics[width=\linewidth]{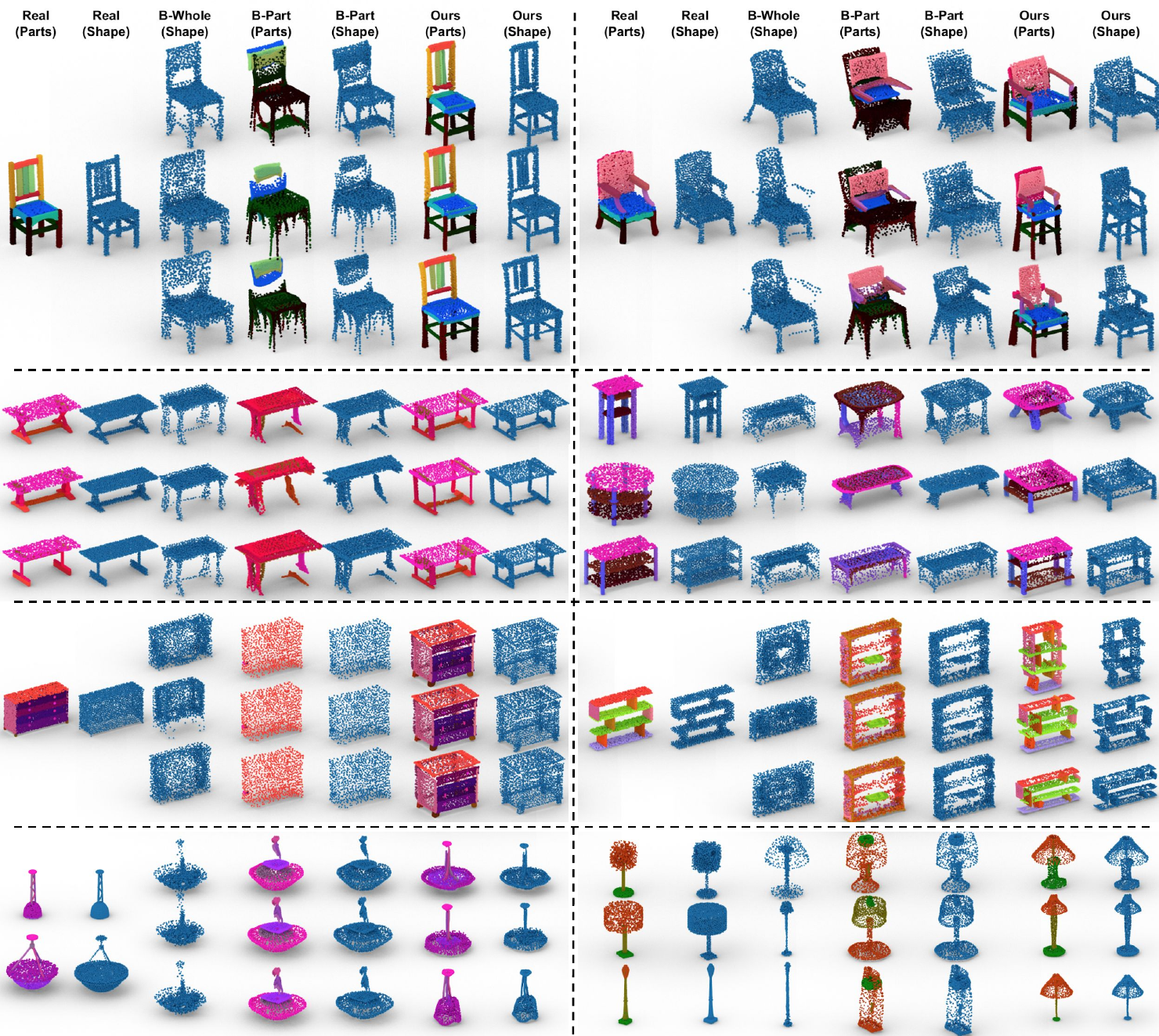}
  \caption{\titlecap{Qualitative Comparisons.}{We show two examples for each of the four categories: chair, table, cabinet and lamp. The leftmost two columns show the real examples illustrating the conditional part tree input (see Figure~\ref{fig:fig4_part_tree} for the input part tree visualization). We show three random real examples unless there are only one or two in the dataset. For our method and \texttt{B-Part} we show both the generated part point clouds with colors and the down-sampled shape point clouds, to fairly compare with \texttt{B-Whole} that only produces shape point clouds.}
  }
  \label{fig:result_compare}
\end{figure}

\subsection{Baselines}
We compare to two vanilla versions of conditional GAN methods as baselines.
\begin{itemize}
    \item \textbf{Whole-shape Vanilla c-GAN (\texttt{B-Whole}):} the method uses a Multiple-layer Perception (MLP) $G_\texttt{baseline}(\mathbf{z}, \mathcal{T})$ as the generator and the holistic shape point cloud discriminator $D^\texttt{whole}_\texttt{baseline}(\mathbf{x},\mathcal{T})$;
    \item \textbf{Part-based Vanilla c-GAN (\texttt{B-Part}):} the method uses exactly the same proposed generator $G(\mathbf{z}, \mathcal{T})$ as in our method and a holistic shape point cloud discriminator $D^\texttt{whole}_\texttt{baseline}(\mathbf{x},\mathcal{T})$.
\end{itemize}
One can think \texttt{B-Part} as an ablated version of our proposed method, without the structural discriminator $D^\texttt{struct}(\mathbf{X})$.
The \texttt{B-Whole} method further removes our part-based generator and generates whole shape point clouds in one shot, similar to previous works~\cite{achlioptas2018learning,valsesia2018learning,shu20193d}.
We implement $D^\texttt{whole}_\texttt{baseline}(\mathbf{x},\mathcal{T})$ similar to $D^\texttt{whole}(\mathbf{x})$ used as part of our discriminator.
It uses a vanilla PointNet~\cite{qi2017pointnet} to extract global geometry features for input point clouds.
Additionally, to make it be aware of the input part tree condition $\mathcal{T}$, we re-purpose the proposed part tree feature encoder network $G_\texttt{enc}$ in our generator to recursively compute a root node feature summarizing the entire part tree structural information.
We make $D^\texttt{whole}_\texttt{baseline}(\mathbf{x},\mathcal{T})$ conditional on the extracted root node feature.
For $G_\texttt{baseline}(\mathbf{z}, \mathcal{T})$, we follow Achlioptas~\etal~\cite{achlioptas2018learning} and design a five-layer MLP with sizes $512$, $512$, $512$, $1024$, $2048\times3$ that finally produces a point cloud of size $2048\times3$. We use leaky \texttt{ReLU} as activation functions except for the final output layer.
We also condition $G_\texttt{baseline}(\mathbf{z}, \mathcal{T})$ on the root feature extracted from the template feature encoder. 

\subsection{Metrics}
We report standard point cloud GAN metrics, including coverage, diversity~\cite{achlioptas2018learning}, and Frech\'et Point-cloud Distance (FPD)~\cite{shu20193d}. Note that coverage and diversity originally measure the distance between shape point clouds, which is, more or less, structure-unaware. We introduce two structure-aware metrics, \emph{part coverage} and \emph{part diversity} adopting the original ones by evaluating the average distance between corresponding parts of the two shapes. In addition, we devise a novel perceptual structure-aware metric \emph{HierInsSeg} that measures the part tree edit distance leveraging deep neural networks. 

\paragraph{Coverage Scores.} Conditioned on every part tree $\mathcal{T}$, the coverage score measures the average distance from each of the real shapes $\mathbf{X}_{i,\texttt{real}}=\left\{X^j_i\mid P^j\in\mathcal{T}_\texttt{leaf}\right\}$ to the closest generated sample in $\left\{\mathbf{X}_{j,\texttt{gen}}\right\}_j$.
\begin{equation}
   \text{Coverage Score}(\mathcal{T})=\frac{1}{\left|\mathcal{X}_\mathcal{T}\right|}\sum_{X_{i,\texttt{real}}\in \mathcal{X}_\mathcal{T}}\left(\min_{j}\texttt{Dist}\left(X_{i,\texttt{real}}, X_{j,\texttt{gen}}\right)\right)
\end{equation}
where $\mathcal{X}_\mathcal{T}$ includes all real data samples $\left\{\mathbf{X}_{i,\texttt{real}}\right\}_i$ that satisfies $\mathcal{T}$.
We randomly generate 100 point cloud shapes $\left\{\mathbf{X}_{j,\texttt{gen}}\right\}_{j=1}^{100}$.

We introduce two variants of function \texttt{Dist} to measure the distance between two sets of part point clouds $\mathbf{X}_{i_1}$ and $\mathbf{X}_{i_2}$.
\begin{align}
    \label{eqn:coverage_scores}
    \texttt{Dist}^\texttt{part}\left(\mathbf{X}_{i_1}, \mathbf{X}_{i_2}\right) =& \frac{1}{\left|\mathcal{T}_\texttt{leaf}\right|}\sum_{(j_1, j_2)\in \mathcal{M}} \texttt{EMD}\left(\mathbf{x}_{i_1}^{j_1}, \mathbf{x}_{i_2}^{j_2}\right)  \nonumber\\
    \texttt{Dist}^\texttt{shape}\left(\mathbf{X}_{i_1}, \mathbf{X}_{i_2}\right) =& \texttt{EMD}\left(\texttt{DownSample}(\mathbf{X}_{i_1}), \texttt{DownSample}(\mathbf{X}_{i_2})\right)
\end{align}
where \texttt{EMD} denotes the Earth Mover Distance~\cite{rubner2000earth,fan2017point} between two point clouds and \texttt{DownSample} is Furthest Point Sampling (FPS).
Here, $\mathcal{M}$ is the solution to a linear sum assignment we compute over two sets of part point clouds $\left\{\mathbf{x}^j_{i_1}\mid P^j\in\mathcal{T}_\texttt{leaf}\right\}$ and $\left\{\mathbf{x}^j_{i_2}\mid P^j\in\mathcal{T}_\texttt{leaf}\right\}$ according to the part tree and part geometry.

We measure \emph{part coverage score} and \emph{shape coverage score} using $\texttt{Dist}^\texttt{part}$ and $\texttt{Dist}^\texttt{shape}$ respectively for every part tree condition $\mathcal{T}$, and finally average over all part trees to obtain the final coverage scores.
The \emph{shape coverage score} measures the holistic shape distance which is less structure-aware, while the \emph{part coverage score} treats all parts equally and is more suitable to evaluate our part-tree conditioned generation task.

\paragraph{Diversity Scores.} A good point cloud GAN should generate shapes with variations.
We generate 10 point clouds for each part tree condition and compute diversity scores under both distance functions $\texttt{Dist}^\texttt{part}$ and $\texttt{Dist}^\texttt{shape}$.
Finally, we report the average \emph{part diversity score} and \emph{shape diversity score} across all part tree conditions.
\begin{equation}
   \label{eqn:diversity_scores}
    \text{Diversity Score}(\mathcal{T})=\frac{1}{100}\sum_{i,j=1}^{10}\left(\texttt{Dist}\left(X_{i,\texttt{gen}}, X_{j,\texttt{gen}}\right)\right)
\end{equation}

\paragraph{Frech\'et Point-cloud Distance.} Shu~\etal~\cite{shu20193d} introduced Frech\'et Point-cloud Distance (FPD) for evaluating the point cloud generation quality,
inspired by the Frech\'et Inception Distance (FID)~\cite{heusel2017gans} commonly used for evaluating 2D image generation quality.
A PointNet~\cite{qi2017pointnet} is trained on ModelNet40~\cite{wu20153d} for 3D shape classification and then FPD computes the real and fake feature distribution distance using the extracted point cloud global features from PointNet.

FPD jointly evaluates the generation quality, diversity and coverage.
It is defined as
\begin{equation}
    \text{Frech{e}t Distance} = ||\mu_r - \mu_g||^2 + \text{Tr}(\Sigma_r + \Sigma_g - 2\left(\Sigma_r\Sigma_g\right)^{1/2}).
\end{equation}
where $\mu$ and $\Sigma$ are the mean vector and the covariance matrix of the features for the real data distribution $r$ and the generated one $g$.
The notation $\text{Tr}$ refers to the matrix trace. 

As most of the part trees in PartNet have only one or few real shapes, we cannot easily compute a stable real data mean $\mu_r$ and covariance matrix $\Sigma_r$ for each part tree, which usually requires hundreds or thousands of data points to compute.
Thus, we have to compute FPD over all part tree conditions by randomly sampling a part tree condition from the dataset and generating one shape point cloud conditioned on it.
In this paper, we generate 10,000 shapes for each evaluation.

\paragraph{HierInsSeg Score.}
We propose a novel \emph{HierInsSeg score}, which is a structural metric that measures how well the generated shape point clouds satisfy the symbolic part tree conditions.
The \emph{HierInsSeg} algorithm $\texttt{Seg}(\mathbf{x})$ performs hierarchical part instance segmentation on the input shape point cloud $\mathbf{x}$ and outputs a symbolic part tree depicting its part structure.
Then we compute a tree-editing distance between this part tree prediction and the part tree used as the generation condition.
We perform a hierarchical Hungarian matching over the two symbolic part trees that matches according to the part semantics and the part subtree structures in a hierarchical top-down fashion.
Any node mismatch in this procedure contributes to the tree difference score and the final tree-editing distance is computed by further divided by the total node count of the input part tree condition.

For each part tree, we conditionally generate 100 shape point clouds and compute the mean tree-editing distance. 
To get the \emph{HierInsSeg} score, we simply average the mean tree-editing distances from all part trees.

Mo~\etal~\cite{Mo_2019_CVPR} proposed a part instance segmentation method that takes as input a point cloud shape and outputs a variable number of disjoint part masks over the point cloud input, each of which represents a part instance.
The method uses PointNet++~\cite{qi2017pointnet++} as a backbone network that extracts per-point features over the input point cloud and then performs a 200-way classification for each point with a SoftMax layer that encourages every point belongs to one mask in the final outputs.
Each of the 200 predicted masks is also associated with a score within $[0, 1]$ indicating its existence.
The existing and non-empty masks correspond to the final part segmentation.
We refer to~\cite{Mo_2019_CVPR} for more details.

We propose our \emph{HierInsSeg} algorithm $\texttt{Seg}(\mathbf{x})$ by adapting~\cite{Mo_2019_CVPR} to a hierarchical setting.
First, we compute the statistics over all training data to obtain the maximal number of parts for each part semantics in the canonical part semantic hierarchy.
Then, we define a maximal instance-level part tree template $\mathcal{T}^\texttt{template}=(\mathcal{T}^\texttt{template}_V, \mathcal{T}^\texttt{template}_E)$ that covers all possible part trees in the training data.
We adapt the same instance segmentation pipeline~\cite{Mo_2019_CVPR} but change the maximal number of output masks from 200 to $\left|\mathcal{T}^\texttt{template}_V\right|$.
Finally, to make sure all children part masks sum up to the parent mask in the part tree template, we define
\begin{equation}
    \mathbf{M}_j=\sum_{(j,k)\in\mathcal{T}^\texttt{template}_E}\mathbf{M}_k, \forall j
\end{equation}
To implement this, for each parent part mask, we perform one SoftMax operation over all children part masks. The root node always has $\mathbf{M}_\texttt{root}=\mathbf{1}$.

\begin{figure}[t]
\centering
  \includegraphics[width=\linewidth]{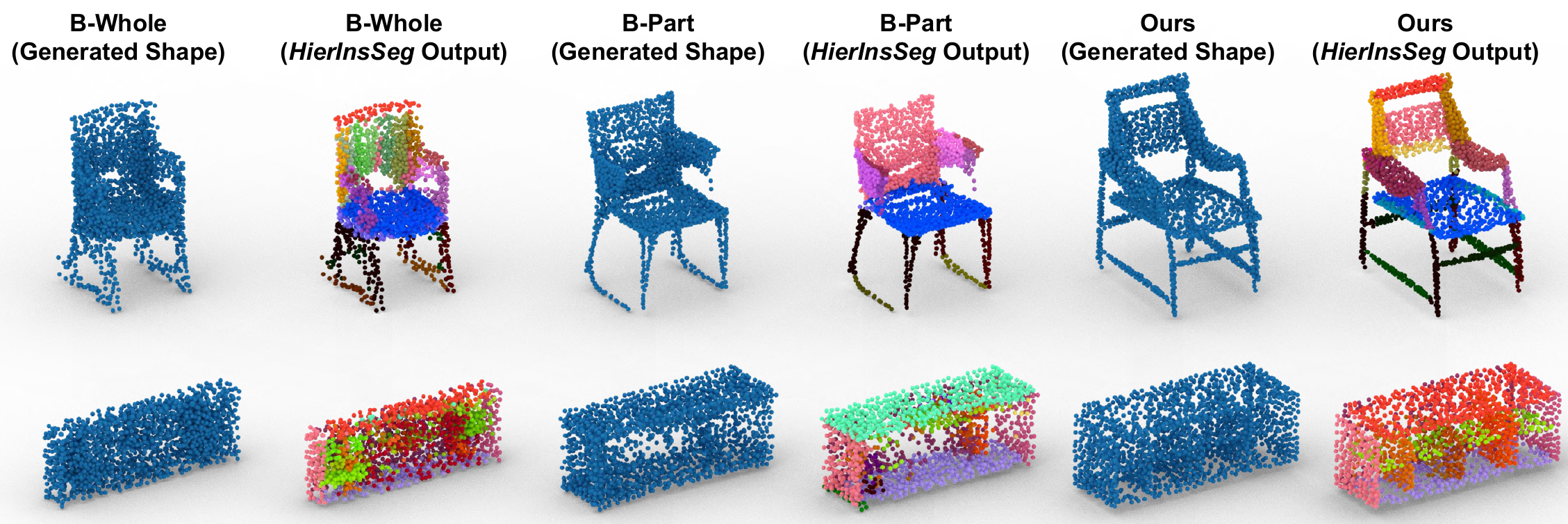}
  \caption{\titlecap{\emph{HierInsSeg} Qualitative Results.}{We show the input generated shape point clouds and the \emph{HierInsSeg} results at the leaf level.}}
  \label{fig:his_results}
\end{figure}

In Table~\ref{result_table} (the \texttt{GT} rows), we present the \emph{HierInsSeg} scores operating on the real shape point clouds to provide a upper bound for the performance.
In Figure~\ref{fig:his_results}, we also show qualitative results for performing the proposed hierarchical instance-level part segmentation over some example generated shapes.
Both quantitative and qualitative results show that the proposed \emph{HierInsSeg} algorithm is effective on judging if the generated shape point cloud satisfies the part tree condition.

\begin{table}[t]
\caption{\titlecap{Quantitative Evaluations.}{We report the quantitative metric scores for our \Ptpc framework and the two vanilla c-GAN baselines. S and P are short for Shape and Part. Cov, Div, HIS are short for \emph{coverage} score, \emph{diversity} score and \emph{HierInsSeg} score. Since the baseline \texttt{B-Whole} does not predict part point clouds, so \emph{part coverage score} and \emph{part diversity score} cannot be defined. We also report the ground-truth \emph{HierInsSeg} scores for each category. The last two rows show the ablation study on chair, where Ours-W is ours without $D^\texttt{whole}$.}} 
\centering
\label{result_table}
{\scriptsize
\begin{tabularx}{\textwidth}{>{\centering}m{0.9cm}|>{\centering}m{1.1cm}|cccccc|cccccc}
    \toprule
    \multirow{2}{*}{} & \multirow{2}{*}{Method} & \multicolumn{6}{c|}{Train} & \multicolumn{6}{c}{Test} \\
    & & S-Cov$\downarrow$ & P-Cov$\downarrow$ & S-Div$\uparrow$ & P-Div$\uparrow$ & FPD$\downarrow$ & HIS$\downarrow$ & S-Cov & P-Cov & S-Div & P-Div & FPD & HIS \\
    \midrule
    \multirow{4}{*}{Chair} 
    & B-Whole & \textbf{0.13} & -- & 0.14 & -- & 7.32 & 0.57 & \textbf{0.13} & -- & 0.13 & -- & 10.88 & 0.57 \\
    & B-Part & 0.14 & 0.41 & 0.14 & 0.06 & 7.17 & 0.58 & 0.15 & 0.41 & 0.14 & 0.06 & 11.10 & 0.58\\
    & Ours & \textbf{0.13} & \textbf{0.06} & \textbf{0.18} & \textbf{0.07} & \textbf{6.64} & \textbf{0.48} & 0.14 & \textbf{0.07} & \textbf{0.18} & \textbf{0.07} & \textbf{10.69} & \textbf{0.48}\\
        & GT & & & & & & \underline{0.30} & & & & & & \underline{0.31} \\
    \midrule
    \multirow{4}{*}{Table} 
    & B-Whole & \textbf{0.19} & -- & 0.14 & -- & 13.02 & 1.04 & \textbf{0.21} & -- & 0.14 & -- & 20.63 &  1.02 \\
    & B-Part & 0.20  &  0.60 & 0.15 & \textbf{0.09} & 6.45 &  1.02  &   \textbf{0.21} & 0.60  & 0.15  & \textbf{0.09} & 16.92 & 0.99 \\
    & Ours & 0.21  & \textbf{0.11}  & \textbf{0.18} & \textbf{0.09}  & \textbf{5.58}  & \textbf{0.93}  &  0.23  & \textbf{0.17} & \textbf{0.17} & \textbf{0.09} & \textbf{15.33} & \textbf{0.89} \\
        & GT & &&&&&\underline{0.62}& & &&&&\underline{0.64} \\
    \midrule
    \multirow{4}{*}{Cabinet} 
    & B-Whole & 0.15 & -- & 0.09 & -- & 16.38 &  0.90 & \textbf{0.17} & -- & \textbf{0.08} & -- & 22.90 & 0.86 \\
    & B-Part & 0.30 & 0.84 & 0.03 & 0.01 & \textbf{3.25} & 0.64 & 0.43 & 0.84 & 0.03 & 0.01 & 24.29 & 0.81 \\
    & Ours & \textbf{0.13} & \textbf{0.08} & \textbf{0.13} & \textbf{0.02} & 4.13 & \textbf{0.52} & 0.24 & \textbf{0.18} & 0.05 & \textbf{0.02} & \textbf{17.73} & \textbf{0.57}\\
        & GT & &&&&&\underline{0.32}& & &&&&\underline{0.35} \\
    \midrule
    \multirow{4}{*}{Lamp} 
    & B-Whole & 0.38 & -- & 0.08 & -- & 17.87 &  1.00 &  \textbf{0.38} & -- & 0.09 & -- & 86.96 & 0.96 \\
    & B-Part & \textbf{0.28} & 0.73 & 0.09 & 0.03 & 6.52 &  0.78 & 0.43 & 0.70 & 0.09 & 0.03 & 94.66 & 0.88 \\
    & Ours & 0.32 & \textbf{0.04} & \textbf{0.11} & \textbf{0.05} & \textbf{5.71} & \textbf{0.68}  & 0.41 & \textbf{0.19}  & \textbf{0.12} & \textbf{0.05} & \textbf{80.55} & \textbf{0.83} \\
        & GT & &&&&&\underline{0.51}& & &&&&\underline{0.57}\\
    \midrule
    Chair & Ours-W & 0.14 & 0.07 & \textbf{0.22} & \textbf{0.08} & 10.60 & 0.51 & 0.15 & \textbf{0.07} & \textbf{0.21} & \textbf{0.08} & 13.52 & 0.49\\
    Abla. & Ours & \textbf{0.13} & \textbf{0.06} & 0.18 & 0.07 & \textbf{6.64} & \textbf{0.48} & \textbf{0.14} & \textbf{0.07} & 0.18 & 0.07 & \textbf{10.69} & \textbf{0.48} \\
    \bottomrule
\end{tabularx}
}
\end{table}

\subsection{Results and Analysis}
We train our proposed \emph{PT2PC} method and the two vanilla c-GAN baselines on the training splits of the four PartNet categories.
The part trees in the test splits are unseen during the training time.
Table~\ref{result_table} summarizes the quantitative evaluations.
Our \emph{HierInsSeg} scores are always the best as we explicitly generate part point clouds and hence render clearer part structures.
Moreover, we win most of the FPD scores, showing that our method can generate realistic point cloud shapes.
Finally, we find that our part-based generative model usually provides higher shape diversity as a result of part compositionality.

Figure~\ref{fig:result_compare} shows qualitative comparisons of our method to the two baseline methods.
One can clearly observe that \texttt{B-Whole} produces holistically reasonable shape geometry but with unclear part structures, which explains why it achieves decent shape coverage scores but fails to match our method under FPD and HIS.
For \texttt{B-Part}, it fails severely for chair, table and cabinet examples that it does not assign clear roles for the parts and the generated part point clouds are overlaid with each other, which explains the high part coverage scores in Table~\ref{result_table}.
Obviously, our method generates shapes with clearer part structures and boundaries. 
We also see a reasonable amount of generation diversity even for part trees with only one real data in PartNet, thanks to the knowledge sharing among similar part tree and sub-tree structures when training a unified and conditional network.
We also conduct an ablation study on chairs where we remove the holistic discriminator $D^\texttt{whole}$.


\begin{table}[t]
\caption{\titlecap{User Study Results on Chair Generation.}{Here we show the average ranking of the three methods. The ranking ranges from 1 (the best) to 3 (the worst). The results on train templates are calculated based on 267 trials while the results on test templates are from the rest 269 trials.}} 
\centering
\label{tab:user_study}
{\scriptsize
\begin{tabularx}{0.7\textwidth}{>{\centering}m{1.2cm}|ccc|ccc}
    \toprule
    \multirow{2}{*}{} &  \multicolumn{3}{c|}{Train} & \multicolumn{3}{c}{Test} \\
   & Structure       & Geometry        & Overall         & Structure       & Geometry        & Overall \\
    \midrule
    B-Whole & 2.39          & 2.07          & 2.22          & 2.40          & 2.10          & 2.21  \\
    B-Part & 2.33          & 2.41          & 2.38          & 2.36          & 2.47          & 2.46   \\
    Ours       & \textbf{1.29} & \textbf{1.51} & \textbf{1.40} & \textbf{1.24} & \textbf{1.43} & \textbf{1.33} \\
    \bottomrule
\end{tabularx}
}
\end{table}

\subsection{User Study}
Although we provide both Euclidean metrics (\ie coverage and diversity scores) and perceptual metrics (\ie FPD and the proposed \emph{HierInsSeg} scores) for evaluating generation quality in Table \ref{result_table}, the true measure of success is human judgement of the generated shapes. For this reason we perform a user study to evaluate the generation quality on chair class.
For each trial, we show users a part tree as the condition, 5 ground truth shapes as references, and 5 randomly generated shape point clouds for each of the three methods. We ask users to rank the methods regarding the following three aspects: 1) structure similarity to the given part tree; 2) geometry plausibility; 3) overall generation quality. For fair comparison, we randomize the order between the methods in all trials and only show the shape point clouds without part labels. In total, we collected 536 valid records from 54 users. 
In Table \ref{tab:user_study}, we report the average ranking of the three methods. Our method significantly outperforms the other two baseline methods on all of the three aspects and on both train and test templates. Please refer to Appendix Sec. A for the user interface and more details.

\subsection{Decoupling Geometry and Structure for PC-GAN}

\begin{wrapfigure}{r}{0.4\textwidth}
\centering
  \includegraphics[width=\linewidth]{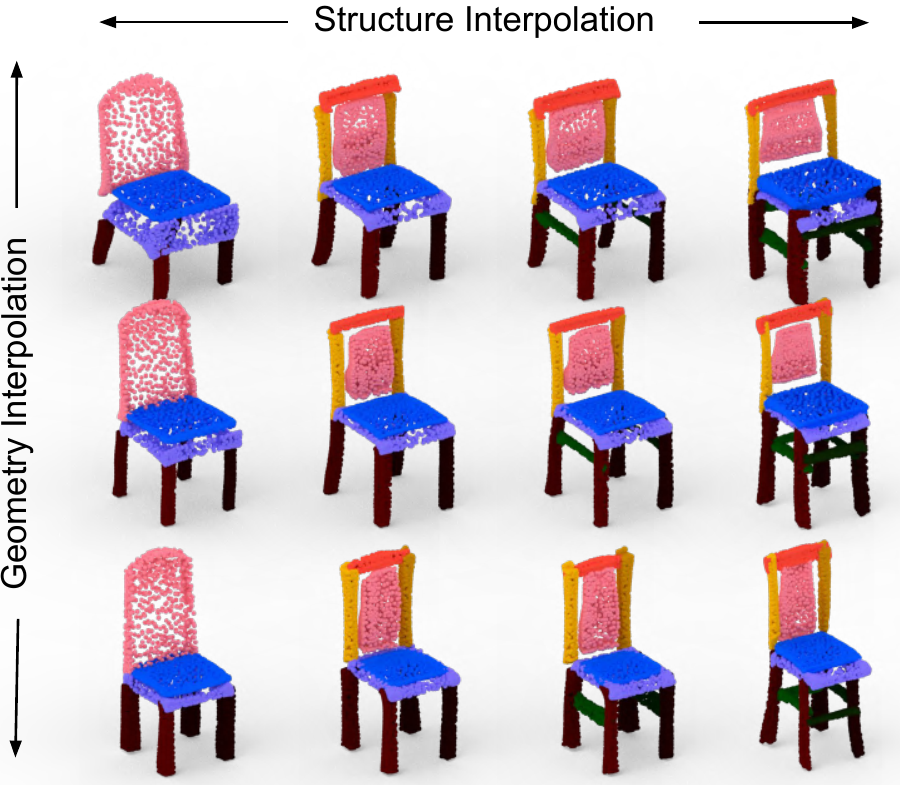}
  \caption{Our approach enables disentanglement of \emph{geometry} and \emph{structure} factors in point cloud generation. Each row shares the same Gaussian noise $\mathbf{z}$ and every column is conditioned on the same part tree input. }
  \label{fig:app}
\end{wrapfigure}
Our proposed \emph{PT2PC} framework enables disentanglement of shape \emph{structure} and \emph{geometry} generation factors.
We demonstrate the capability of exploring structure-preserving geometry variation and geometry-preserving structure variation using our method.
Conditioned on the same symbolic part tree, our network is able to generate shape point clouds with \emph{geometric} variations by simply changing the Gaussian random noise $\mathbf{z}$.
On the other hand, if we fix the same noise $\mathbf{z}$, conditioned on different input part trees, we observe that \emph{PT2PC} is able to produce \emph{geometrically} similar but \emph{structurally} different shapes.
Figure~\ref{fig:app} shows the generated shape point clouds $\left\{\mathbf{x}_{i,j}=G\left(\mathbf{z}_i, \mathcal{T}_j\right)\right\}$ from a set of Gaussian noises $\left\{\mathbf{z}_i\right\}_i$ and a set of part trees $\left\{\mathcal{T}_j\right\}_j$.
Each row shows shape structural interpolation results while sharing similar shape geometry, and every column presents geometric interpolation results conditioned on the same part tree structure.


\section{Conclusion}
\label{sec:conclusion}
We have proposed \emph{PT2PC}, a conditional generative adversarial network (c-GAN) that generates point cloud shapes given a symbolic part-tree condition.
The part tree input specifies a hierarchy of semantic part instances with their parent-children relationships.
Extensive experiments and user study show our superior performance compared to two baseline c-GAN methods.
We also propose a novel metric \emph{HierInsSeg} to evaluate if the generated shape point clouds satisfy the part tree conditions.
Future works may study incorporating more part relationships and extrapolating our method to unseen categories.

\section*{Acknowledgments}
This research was supported by a Vannevar Bush Faculty Fellowship, grants from the Samsung GRO program and the Stanford SAIL Toyota Research Center, and gifts from Autodesk and Adobe.

%
%
\bibliographystyle{splncs04}
\bibliography{egbib}

\section*{Appendix}
\subsection*{A. More Details on the User Study}
\label{sec:supp_ui}
We show our user study interface in Figure~\ref{fig:ui}.
We ask the users to rank three algorithms from three aspects: part structure, geometry, overall.

\subsection*{B. More Qualitative Results}
We present more qualitative results in Figure~\ref{fig:gallery}.
Given the symbolic part trees as conditions, we show multiple diverse point clouds generated by our method.

\subsection*{C. Mesh Generation Results}
\begin{figure}[h]
\centering
  \includegraphics[width=\linewidth]{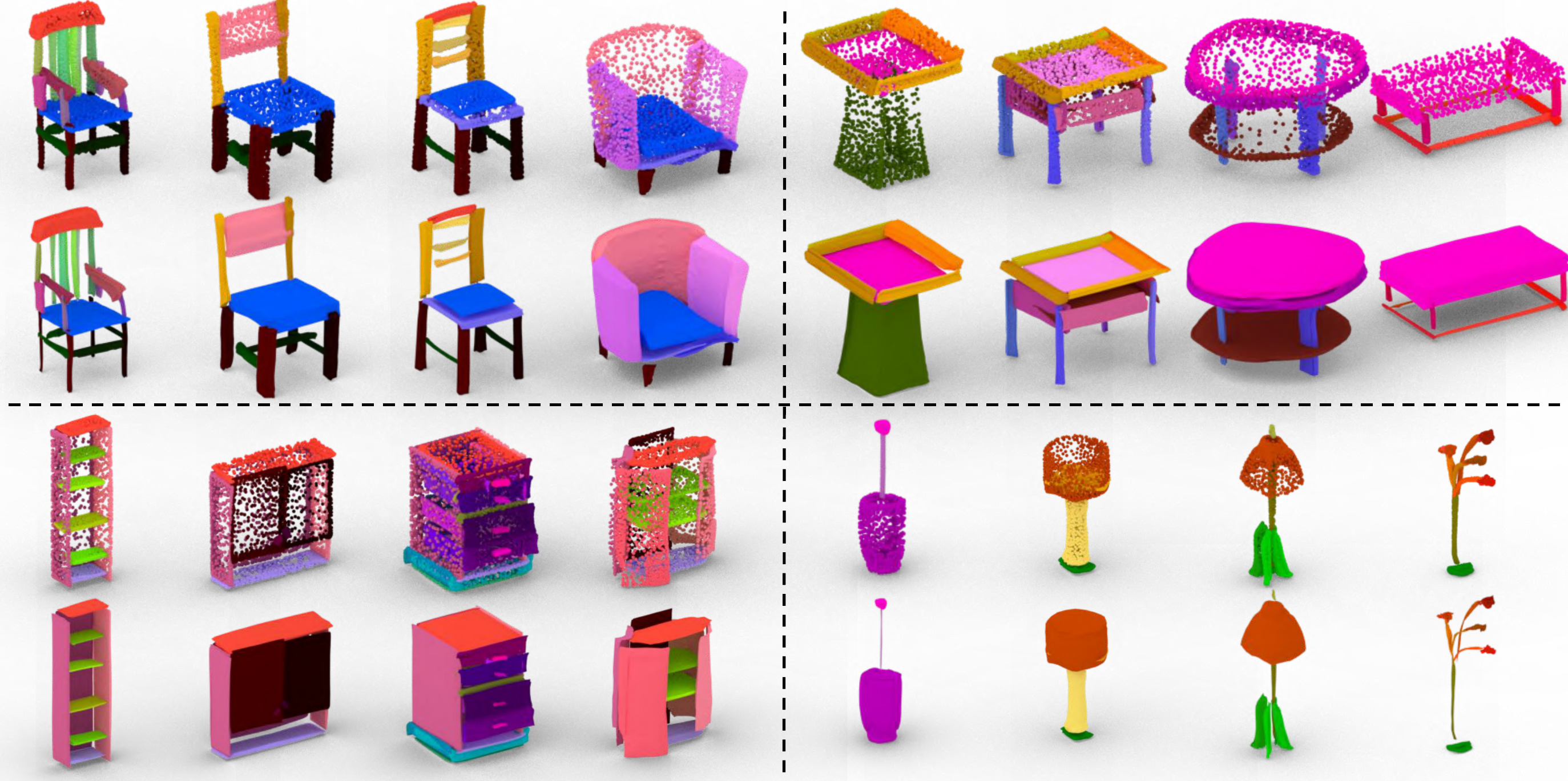}
  \caption{\titlecap{Mesh Generation Results.}{The top rows show the generated shape point clouds and the bottom rows show the corresponding generated mesh results.}}
  \label{fig:mesh_gen}
\end{figure}

Since our method deforms a point cloud sampled from a unit cube mesh for each leaf-node part geometry, we naturally obtain the mesh generation results as well.
Figure~\ref{fig:mesh_gen} shows some results.
Since the goal of this work is primarily for point cloud generation, we do not explicitly optimize for the mesh generation results.
However, we observe reasonable mesh generation results learned by our method.

\subsection*{D. Failure Cases and Future Works}
\begin{figure}[t]
\centering
  \includegraphics[width=0.7\linewidth]{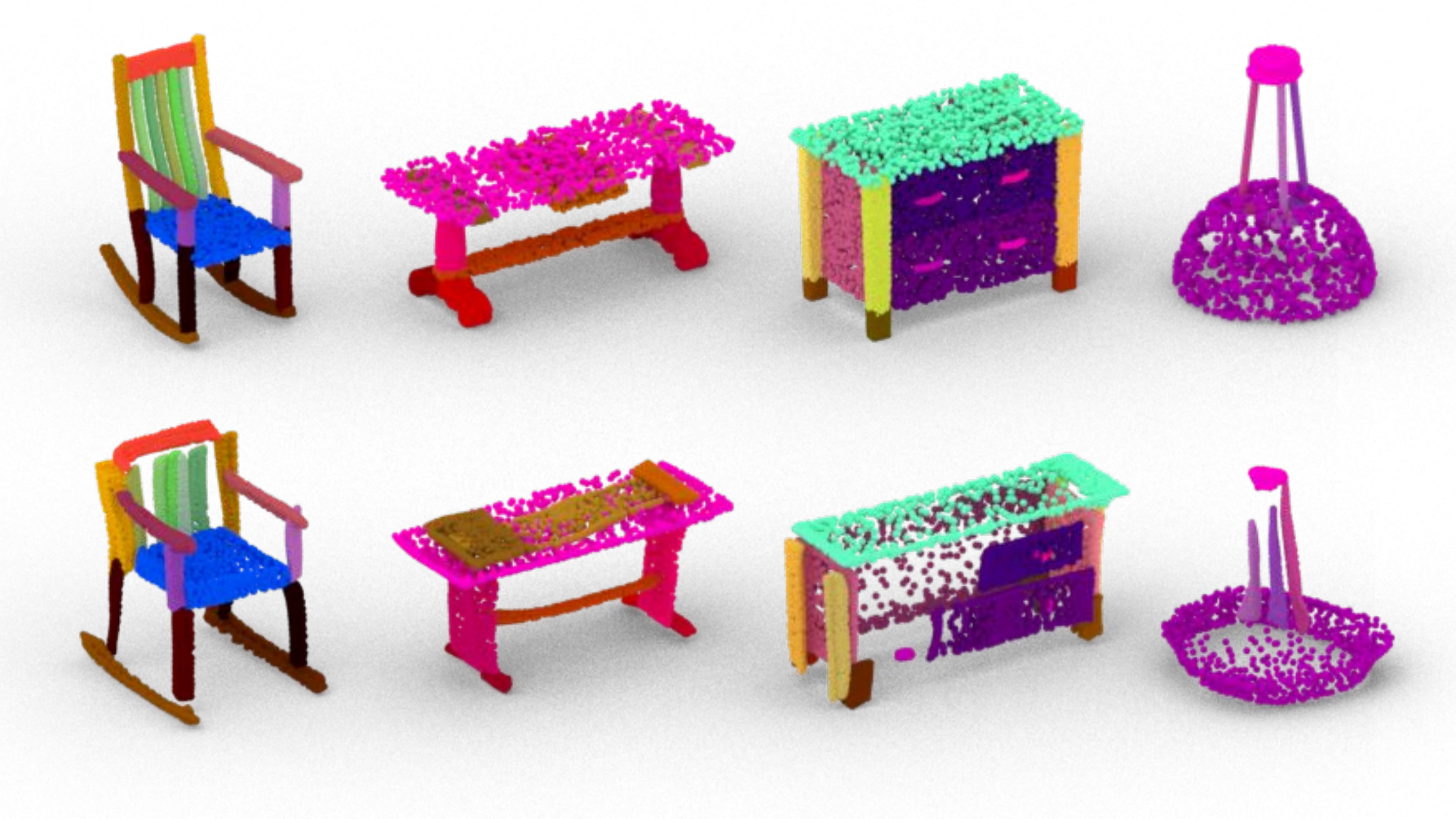}
  \caption{\titlecap{failure cases.}{The top row shows the real shapes and the bottom row presents our generated point clouds.}}
  \label{fig:failure}
\end{figure}

\begin{figure}[t]
\centering
  \includegraphics[width=\linewidth]{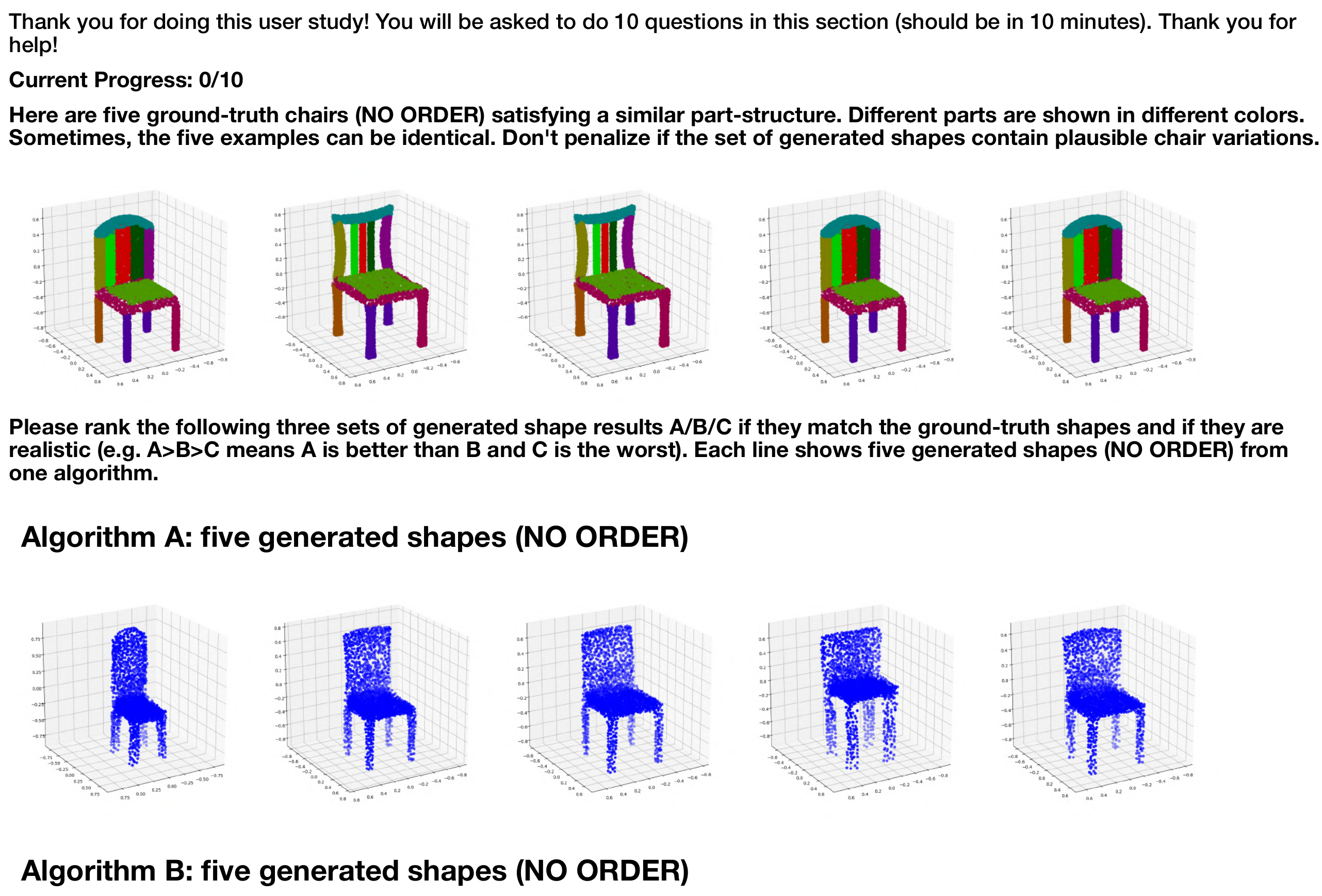}
  \includegraphics[width=\linewidth]{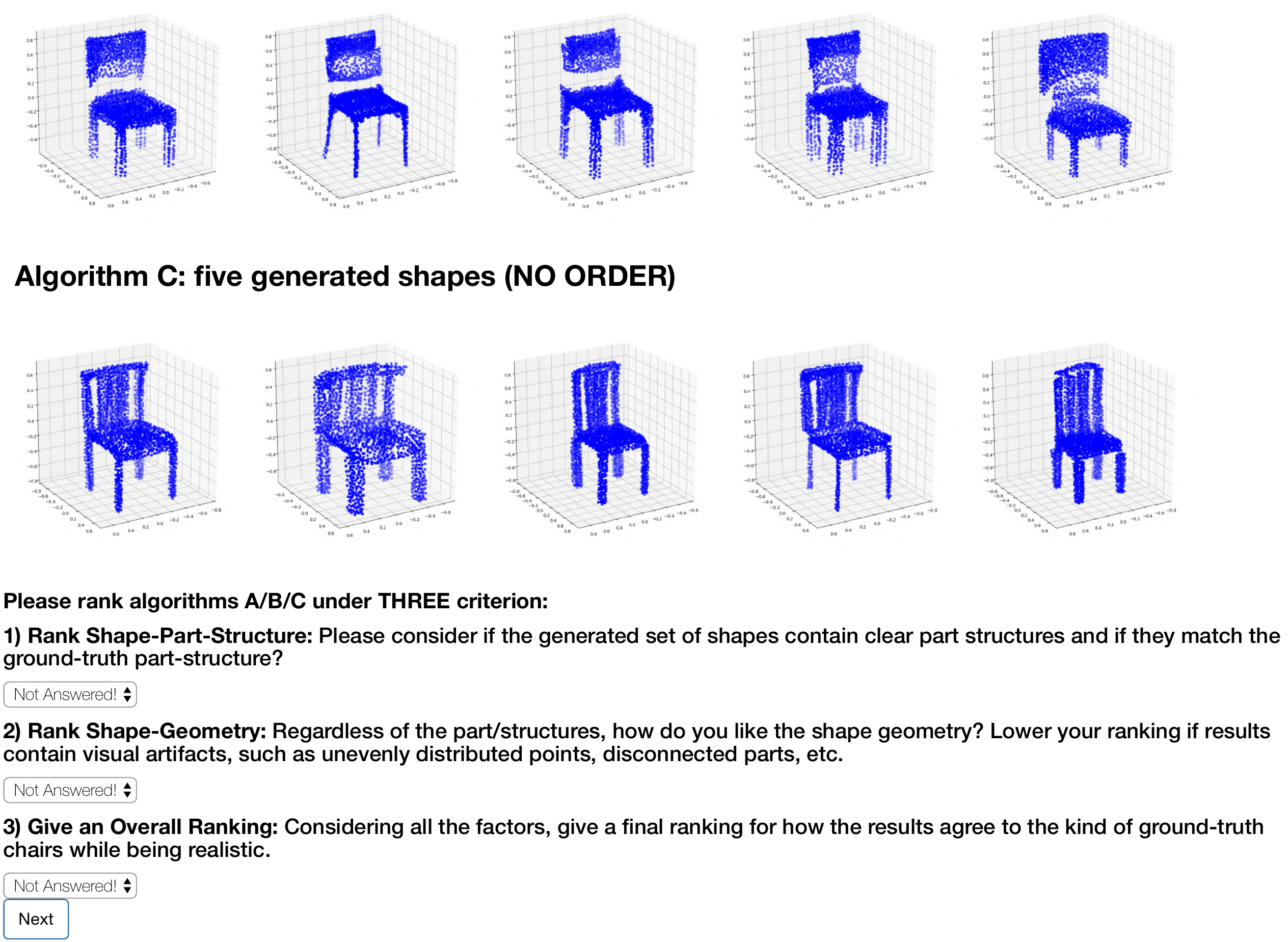}
  \caption{\textbf{User Study Interface.}}
  \label{fig:ui}
\end{figure}

\begin{figure}[t]
\centering
  \includegraphics[width=\linewidth]{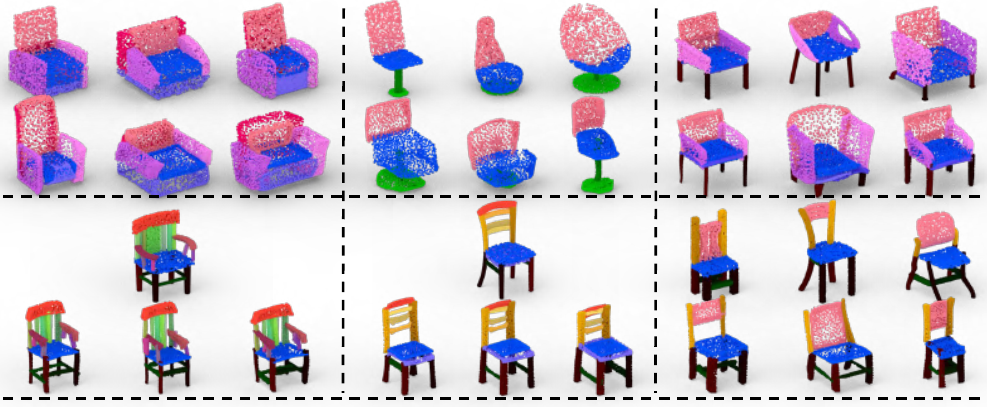}
  \includegraphics[width=\linewidth]{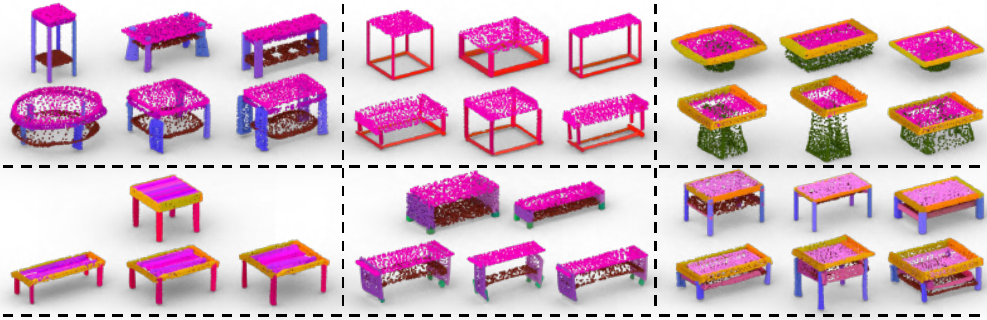}
  \includegraphics[width=\linewidth]{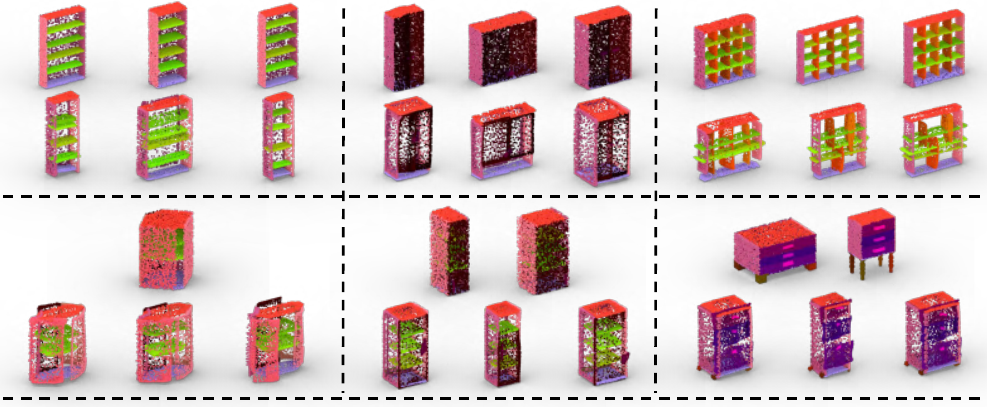}
  \includegraphics[width=\linewidth]{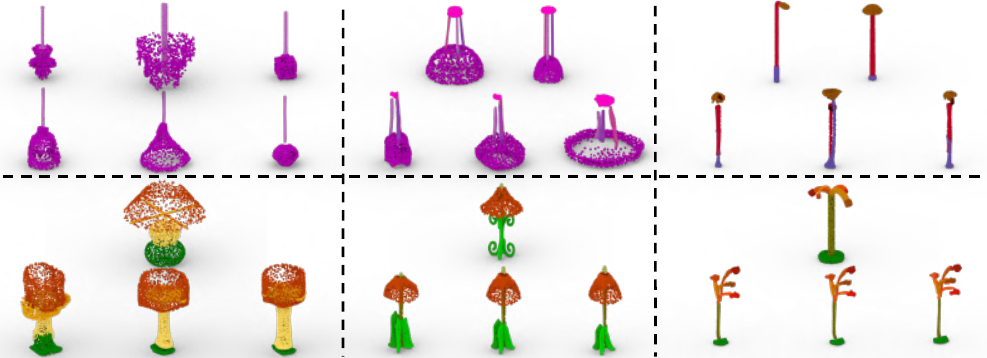}
  \caption{\titlecap{Additional qualitative results.}{We show six more results for each of the four categories. For each block, the top row shows the real shapes and the bottom row shows our generated results.}}
  \label{fig:gallery}
\end{figure}

Figure~\ref{fig:failure} presents common failure cases we observe.
For the chair example, the back slats are not well aligned with each other and are unevenly distributed spatially. 
For the table example, the connecting parts between legs and surface extrude outside the table surface.
In the cabinet example, the four drawers overlap with each other as the network does not assign clear roles for them.
The lamp example has the disconnection problem between the rope and the base on the ceiling.
All these cases indicate that future works should study how to better model part relationships and physical constraints.

\subsection*{E. Part Tree Visualization for Figure~\ref{fig:result_compare}}
\begin{figure}[t]
\centering
  \includegraphics[width=\linewidth]{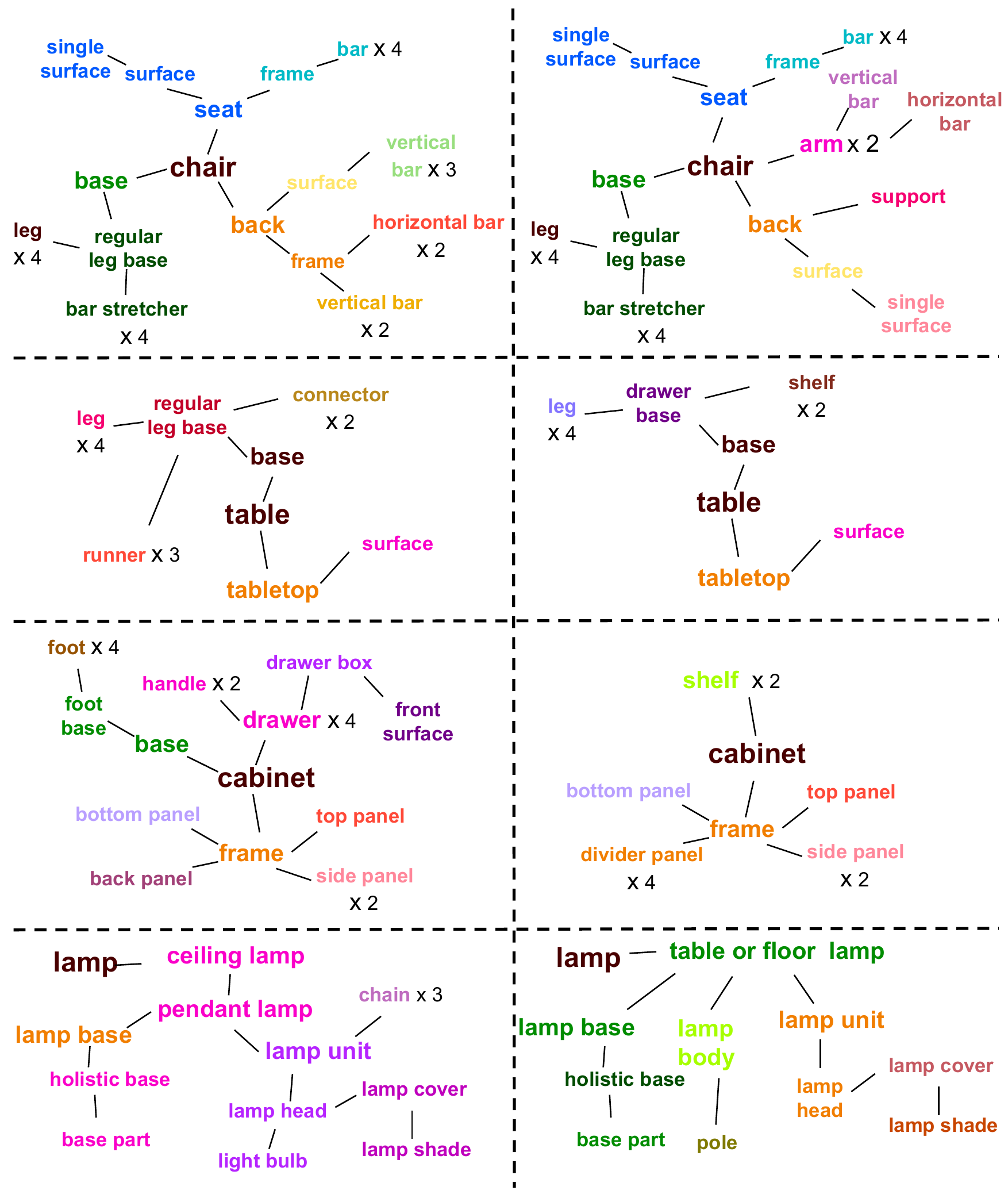}
  \caption{\textbf{Visualization for the Part Tree Conditions for Figure~\ref{fig:result_compare}.} Here we show the eight part tree conditional inputs used for generating the point cloud shapes in Figure~\ref{fig:result_compare}.}
  \label{fig:fig4_part_tree}
\end{figure}

Figure~\ref{fig:fig4_part_tree} shows the eight part tree conditional inputs used for generating the point cloud shapes in Figure~\ref{fig:result_compare}.

\end{document}